\documentclass{article}

\PassOptionsToPackage{numbers,compress}{natbib}

\usepackage[final]{nips_2017}

\usepackage[utf8]{inputenc} 
\usepackage[T1]{fontenc}    
\usepackage{hyperref}       
\usepackage{url}            
\usepackage{booktabs}       
\usepackage{amsfonts}       
\usepackage{nicefrac}       
\usepackage{microtype}      
\usepackage{color}
\usepackage{graphicx}
\usepackage{amsmath}
\usepackage[algoruled,linesnumbered,noend]{algorithm2e}

\newcommand{\argmax}[1]{\underset{#1}{\operatorname{arg}\,\operatorname{max}}\;}

\title{Explaining Transition Systems\\ through Program Induction} 

\author{
  Svetlin Penkov \\
  School of Informatics \\
  The University of Edinburgh \\
  \texttt{sv.penkov@ed.ac.uk} \\
  \And
  Subramanian Ramamoorthy \\
  School of Informatics \\
  The University of Edinburgh \\
  \texttt{s.ramamoorthy@ed.ac.uk} \\
}

\begin{document}

\maketitle

\begin{abstract}
Explaining and reasoning about processes which underlie observed black-box phenomena enables the discovery of causal mechanisms, derivation of suitable abstract representations and the formulation of more robust predictions. We propose to learn high level functional programs in order to represent abstract models which capture the invariant structure in the observed data. We introduce the $\pi$-machine (program-induction machine) -- an architecture able to induce interpretable LISP-like programs from observed data traces. We propose an optimisation procedure for program learning based on backpropagation, gradient descent and A* search. We apply the proposed method to three problems: system identification of dynamical systems, explaining the behaviour of a DQN agent and learning by demonstration in a human-robot interaction scenario. Our experimental results show that the $\pi$-machine can efficiently induce interpretable programs from individual data traces.

\end{abstract}

\section{Introduction}
Learning models of transition systems has been a core concern within machine learning, with applications ranging from system identification of dynamical systems \citep{schmidt2009} and inference of human choice behaviour \citep{Glimcher2011, brendel2011} to reverse engineering the behaviour of a device or computer program from observations and traces \citep{Vaandrager2017}. With the increasing use of these learnt models in the inner loops of decision making systems, e.g., in robotics and human-machine interfaces, it has become necessary to ensure not only that these models are accurate predictors of behaviour, but also that their causal mechanisms are exposed to the system designer in a more interpretable manner. There is also the need to explain the model in terms of counterfactual reasoning \citep{bottou2013}, e.g., what would we expect the system to do if a certain variable were changed or removed, or model checking \citep{Baier2008} of longer term properties including safety and large deviations in performance. We address these needs through a program induction based framework for explainability.

We propose to learn high level functional programs in order to represent abstract models which capture the invariant structure in the observed data. Recent works have demonstrated the usefulness of program representations in capturing human-like concepts \citep{lake2015}. Used in this way, program-based representations boost generalisation and enable one-shot learning. Also, and arguably more importantly, they are significantly more amenable to model checking and human interpretability. 

\begin{figure}
	\centering
    	\includegraphics[width=1.0\textwidth]{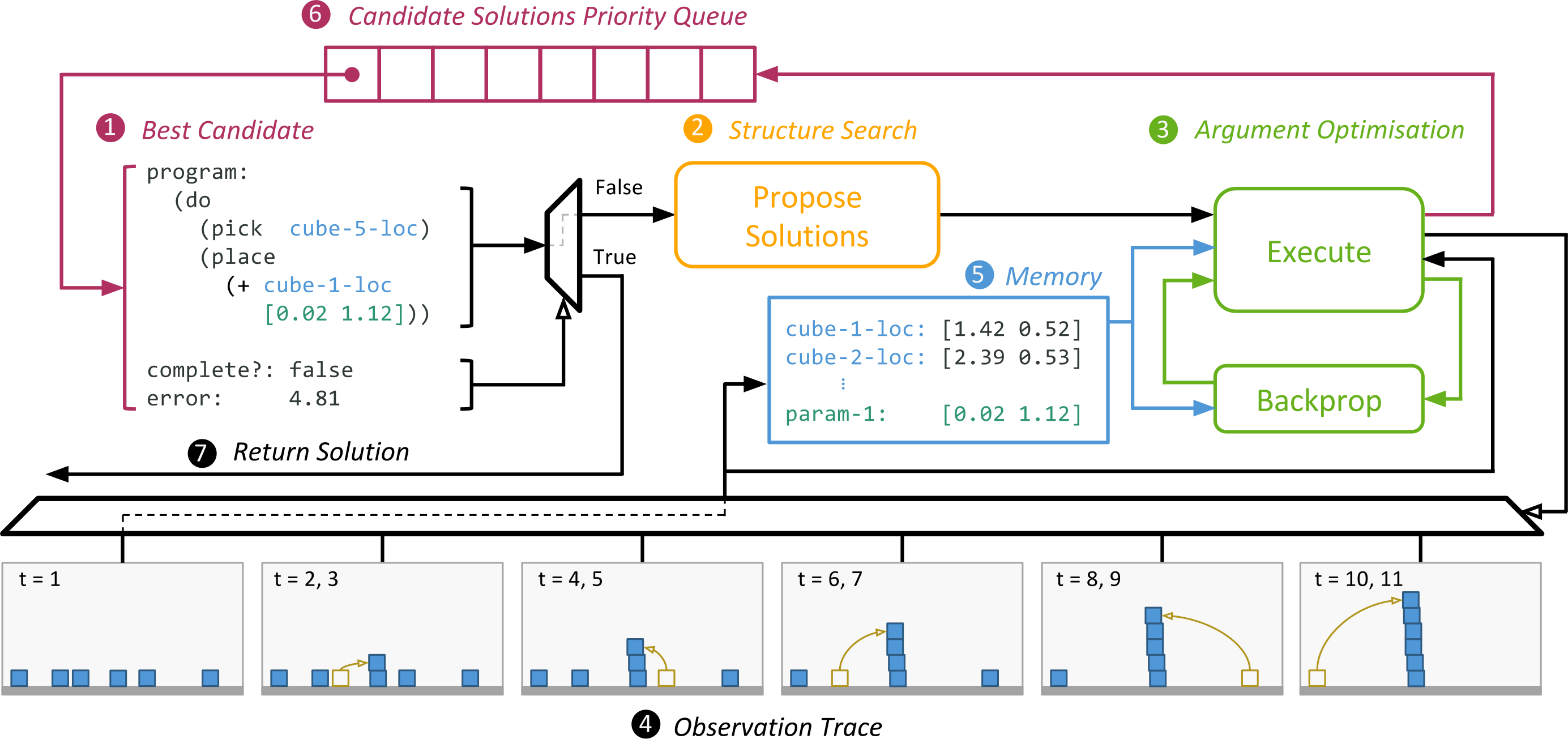}
    \caption{Overall architecture of the $\pi$-machine. The current best candidate solution (1) is used to propose new, structurally more complex candidates (2). Each one of the new candidate programs $\rho$ is optimised (3) through gradient descent by comparing its execution trace to the observation trace (4). The observation trace in this case is a demonstration of a tower building task. During execution, the program has access to memory (5) which stores both state variables and induced parameters. All new candidate programs are scored based on their performance and complexity and are inserted in the candidate solutions priority queue (6). Once the execution trace of a candidate matches the observation trace the final solution is returned (7).}
    \label{fig:pim_overview}
\end{figure}

In this paper, we introduce the $\pi$-machine (program-induction machine), an architecture which is able to induce LISP-like programs from observed transition system data traces in order to explain various phenomena. Inspired by differentiable neural computers \citep{graves2014, graves2016}, the $\pi$-machine, as shown in Figure \ref{fig:pim_overview}, is composed of a memory unit and a controller capable of learning programs from data by exploiting the scalability of stochastic gradient descent. However, the final program obtained after training is not an opaque object encoded in the weights of a controller neural network, but a LISP-like program which provides a rigorous and interpretable description of the observed phenomenon. A key feature of our approach is that we allow the user to specify the properties they are interested in understanding as well as the context in which the data is to be explained - by providing a set of predicates of interest. By exploiting the equivalence between computational graphs and functional programs we describe a hybrid optimisation procedure based on backpropagation, gradient descent, and A* search which is used to induce programs from data traces. 

We evaluate the performance of the $\pi$-machine on three different problems. Firstly, we apply it to data from physics experiments and show that it is able to induce programs which represent fundamental laws of physics. The learning procedure has access to relevant variables, but it does not have any other prior knowledge regarding physical laws which it has discovered in the same sense as in \citep{schmidt2009} although far more computationally tractably. We then study the use of the proposed procedure in explaining control policies learnt by a deep Q-network (DQN). Starting from behaviour traces of a reinforcement learning agent that has learnt to play the game of Pong, we demonstrate how the $\pi$-machine learns a functional program to describe that policy.  Finally, we consider the domain of learning by demonstration in human-robot interaction, where the $\pi$-machine successfully induces programs which capture the structure of the human demonstration. In this domain, the learnt program plays a key role in enabling the grounding of abstract knowledge (e.g., in natural language commands) in the embodied sensory signals that robots actually work with, as in \citep{penkov2017}.

\section{Related work}
\paragraph{Explainability and interpretability.}
The immense success of deep neural network based learning systems and their rapid adoption in numerous real world application domains has renewed interest in the interpretability and explainability of learnt models \footnote{See, for instance, the end user concerns that motivate the DARPA Explainable AI Programme: \url{http://www.darpa.mil/program/explainable-artificial-intelligence}}. There is recognition that Bayesian rule lists \citep{letham2015, yang2016}, decision trees and probabilistic graphical models are interpretable to the extent that they impose strong structural constraints on models of the observed data and allow for various types of queries, including introspective and counterfactual ones. In contrast, deep learning models usually are trained `per query' and have numerous parameters that could be hard to interpret. \citet{zeiler2014} introduced deconvolutional networks in order to visualise the layers of convolutional networks and provide a more intuitive understanding of why they perform well. Similar approaches can be seen in \citep{bojarski2017,kim2017}, in the context of autonomous driving. \citet{zahavy2016} describe Semi-Aggregated Markov Decision Process (SAMDP) in order to analyse and understand the behaviour of a DQN based agent. Methods for textual rationalisation of the predictions made by deep models have also been proposed \citep{harrison2017,hendricks2016,lei2016}. While all of these works provide useful direction, we need many more methods, especially generic approaches that need not be hand-crafted to explain specific aspects of individual models. In this sense, we follow the model-agnostic explanation approach of \citet{ribeiro2016}, who provide ``textual or visual artefacts'' explaining the prediction of any classifier by treating it as a black-box. Similarly to the way in which \citep{ribeiro2016} utilise local classifiers composed together to explain a more complex model, we present an approach to incrementally constructing functional programs that explain a complex transition system from more localised predicates of interest to the user.

The $\pi$-machine treats the process which has generated the observed data as a black-box and attempts to induce LISP-like program which can be interpreted and used to explain the data. We show that the proposed method can be applied both to introspection of machine learning models and to the broader context of autonomous agents.

\paragraph{Program learning and synthesis.}
Program learning and synthesis has a long history, with the long-standing challenge being the high complexity deriving from the immense search space. Following classic and pioneering work such as by \citet{Shapiro1983} who used inductive inference in a logic programming setting, others have developed methods based on a variety of methods ranging from SAT solvers \citep{solar2006} to genetic algorithms \citep{schmidt2009}, which tend to scale poorly hence often become restricted to a narrow class of programs. Recently, deep neural networks have been augmented with a memory unit resulting in models similar to the original von Neumann architecture. These models can induce programs through stochastic gradient descent by optimising performance on input/output examples \citep{graves2014,graves2016,grefenstette2015} or synthetic execution traces \citep{reed2015,cai2017,ling2017}. Programs induced with such neural architectures are encoded in the parameters of the controller network and are, in general, not easily interpretable (particularly from the point of view of being able to ask counterfactual questions or performing model checking). Interestingly, paradigms from functional programming such as pure functions and immutable data structures have been shown to improve performance of neural interpreters \citep{Keser17}. Another approach is to directly generate the source code of the output program which yields consistent high level programs. Usually, these types of approaches require large amounts of labelled data - either program input/output examples \citep{devlin2017,balog2016} or input paired with the desired output program code \citep{yin2017}.

Determining how many input/output examples or execution traces are required in order to generalise well is still an open research problem. However, in this paper, we focus attention more on the explanatory power afforded by programs rather than on the broader problems of generalisation in the space of programs. While these characteristics are of course related, we take a view similar to that of \cite{ribeiro2016}, arguing that it is possible to build from locally valid program fragments which provide useful insight into the black-box processes generating the data. By combining gradient descent and A* search the $\pi$-machine is able to learn informative and interpretable high-level LISP-like programs, even just from a single observation trace. 

\section{Problem definition}
Consider the labelled transition system $\Omega(\mathcal{S}, \mathcal{A}, \delta)$ where $\mathcal{S}$ is a non-empty set of states, $\mathcal{A}$ is a non-empty set of actions, each parametrised by $\theta \in \mathbb{R}^D$, and $\delta: \mathcal{S} \times \mathcal{A} \rightarrow \mathcal{S}$ is the state transition function. We define an observation trace $\mathcal{T}$ as a sequence of observed state-action pairs $(s_t, a_t(\theta_t)) \in \mathcal{S} \times \mathcal{A}$ generated by the recursive relationship $s_{t+1} = \delta \left(s_t, a_t\right(\theta_t))$ for $1 \leq t \leq T$. We are interested in inducing a LISP-like functional program $\rho$ which when executed by an abstract machine is mapped to an execution trace $\mathcal{T}_\rho$ such that $\mathcal{T}_\rho$ and $\mathcal{T}$ are equivalent according to an input specification. 

We represent the abstract machine as another labelled transition system $\Pi(\mathcal{M}, \mathcal{I}, \varepsilon)$ where $\mathcal{M}$ is the set of possible memory state configurations, $\mathcal{I}$ is the set of supported instructions and $\varepsilon: \mathcal{M} \times \mathcal{I} \rightarrow \mathcal{M}$ specifies the effect of each instruction. We consider two types of instructions -- primitive actions which emulate the execution of $a \in \mathcal{A}$ or arithmetic functions $f \in \mathcal{F}$ such that $\mathcal{I} = \mathcal{A} \cup \mathcal{F}$. Furthermore, a set of observed state variables $\mathcal{M}_v \subseteq \mathcal{S}$, which vary over time, are stored in memory together with a set of induced free parameters $\mathcal{M}_p$. The variables in $\mathcal{M}_v$ form a context which the program will be built on. A custom  detector $\mathcal{D}_v$, operating on the raw data stream, could be provided for each variable, thus enabling the user to make queries with respect to different contexts and property specifications.

The execution of a program containing primitive actions results in a sequence of actions. Therefore, we represent a program $\rho$ as a function which maps a set of input variables $x_v \subset \mathcal{M}_v$ and a set of free parameters $x_p \subset \mathcal{M}_p$ to a finite sequence of actions $\hat{a}_1(\hat{\theta_1}), \ldots \hat{a}_{T'}(\hat{\theta_{T'}})$. We are interested in inducing a program which minimises the total error between the executed and the observed actions:
\begin{equation}
L(\rho) = \sum_{t=1}^{\min(T, T')} \sigma_{act}(\hat{a}_t, \hat{\theta_t}, a_t, \theta_t) + \sigma_{len}(T, T')
\label{eq:loss}
\end{equation}
The error function $\sigma_{act}$ determines the difference between two actions, while $\sigma_{len}$ compares the lengths of the generated and observed action traces.  By providing the error functions $\sigma_{act}$ and $\sigma_{len}$ one can target different aspects of the observation trace to be explained as they specify when two action traces are equivalent.

\section{Method}

The proposed program induction procedure is based on two major steps. Firstly, we explain how a given functional program can be optimised such that the loss $L(\rho)$ is minimised. Secondly, we explain how the space of possible program structures can be searched efficiently by utilising gradient information. An architectural overview of the $\pi$-machine is provided in Figure \ref{fig:pim_overview}. 

\subsection{Program optimisation}

\begin{figure}[h]
	\centering
    	\includegraphics[width=0.6\textwidth]{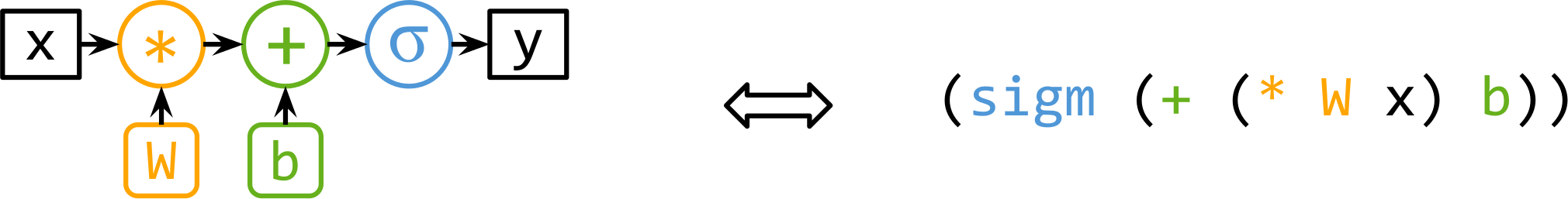}
    \caption{Logistic regression expressed as: computational graph (left); functional program (right).}
    \label{fig:equivalence}
\end{figure}

\paragraph{Functional programs as computational graphs.}
Neural networks are naturally expressed as computational graphs which are the most fundamental abstraction in computational deep learning frameworks \cite{tokui2015,bergstra2010,abadi2016}. Optimisation within a computational graph is usually performed by pushing the input through the entire graph in order to calculate the output (forward pass) and then backpropagating the error signal to update each parameter (backward pass). A key observation for the development of the $\pi$-machine is that computational graphs and functional programs are equivalent as both describe arbitrary compositions of pure functions applied to input data. For example, Figure \ref{fig:equivalence} shows how a logistic regression classifier can be represented both as a computational graph and as a functional program. Therefore, similarly to a computational graph, a functional program can also be optimised by executing the program (forward pass), measuring the error signal and then performing backpropagation to update the program (backward pass). 

\paragraph{Forward pass.}
When a program is executed it is interpreted to a sequence of instructions $i_1, \ldots, i_n \in \mathcal{I}$ which are executed by recursively calling  $\varepsilon(\ldots\varepsilon(\varepsilon(\mathcal{M}_1, i_1), i_2)\ldots, i_n)$. $\mathcal{M}_1$ is the initial memory state initialised with the observed variables from $s_1$ and any induced parameters. The $\pi$-machine keeps a time counter $t$ which is initialised to 1 and is automatically incremented whenever a primitive action instruction is executed. If the instruction $i_k$ is a primitive action, $i_k \in \mathcal{A}$, then the $\pi$-machine automatically sets $\hat{a}_t = i_k$ and invokes the error function $\sigma_{act}(\hat{a}_t, \hat{\theta_t}, a_t, \theta_t)$, where $\hat{\theta_t}$ has been calculated by previous instructions. If the error is above a certain threshold $e_{max}$ the program execution is terminated and the backward pass is initiated. Otherwise, the time counter is incremented and the values of the variables in $\mathcal{M}_v$ are automatically updated to the new observed state. Essentially, the $\pi$-machine simulates the execution of each action reflecting any changes it has caused in the observed state. Alternatively, if the currently executed instruction $i_k$ is a function, $i_k \in \mathcal{F}$, then the resulting value is calculated and $i_k$, together with its arguments, is added to a detailed call trace $\chi$ maintained by the $\pi$-machine. Importantly, each function argument is either a parameter or a variable read from memory at time $t$ or the result of another function. All this information is kept in $\chi$ which eventually contains the computational tree for the entire program.

\paragraph{Backward pass.}
The gradients of the loss function $L(\rho)$ with respect to the program inputs $x_v$ and $x_p$ are required to perform a gradient descent step. Crucially, programs executed by the $\pi$-machine are automatically differentiated. The $\pi$-machine performs reverse-mode automatic differentiation, similarly to Autograd \cite{maclaurin2015}, by traversing the call trace $\chi$, and post-multiplying Jacobian matrices. We assume that the Jacobian matrix with respect to every input argument of any function $f \in \mathcal{F}$ or any specified error function $\sigma_{act}$ is known a priori. Let $f \in \mathcal{F}$ be a function whose output needs to be differentiated with respect to the input arguments. There are three types of derivatives, which need to be considered in order to traverse backwards the entire tree of computations:
\begin{enumerate}
\item Let $g \in \mathcal{F}$, then $\frac{\partial f}{\partial g}$ is the Jacobian matrix of $f$ with respect to the output of $g$ and can be directly calculated. 
\item Let $p \in x_p$, then the gradient $\frac{\partial f}{\partial p}$ is calculated by multiplying the corresponding Jacobian matrix of $f$ with the value of $p$. 
\item Let $v \in x_v$, then the gradient $\frac{\partial f}{\partial v}\Bigr|_{t=t_{r}}$ is calculated by multiplying the corresponding Jacobian of $f$ with the value of the variable at the time it was read from memory $t_{r}$.
\end{enumerate}

\paragraph{Gradient descent step.}
Once the gradient $\nabla_p L(\rho)$ of the loss function with respect to each input parameter $p \in x_p$ is calculated we utilise AdaGrad \cite{duchi2011} to update the values of all parameters after each program execution. The gradient $\nabla_v L(\rho)$ with respect to each input variable $v \in x_v$ is also available. However, a variable cannot be simply updated in the direction of the gradient as it represents a symbol, not just a value. Variables can only take values from memory which is automatically updated according to the observation trace during execution. Nevertheless, the gradient provides important information about the direction of change which we utilise to find variables that minimise the loss. Whenever the memory state is automatically updated, a KD-tree is built for each type of variable stored in memory. We assume that the variables in memory are real vectors with different length. So, we represent the KD-tree which stores all $D$-dimensional variables in memory at time $t$ as $\mathcal{K}^D_t$. If a $d$-dimensional variable $v$ is to be optimised it is replaced with a temporary parameter $p_{temp}$ initialised with $v_t$ which is the value of $v$ read from memory at the respective time step $t$. The temporary parameter $p_{temp}$ is also updated with AdaGrad \cite{duchi2011}. After each descent step, the nearest neighbour of the updated value $p'_{temp}$ is determined by querying the KD-tree with $\mathcal{K}^d_t(p'_{temp})$. If the result of the query is a different $d$-dimensional variable $u$ then the temporary parameter is immediately set to $p_{temp} = u_t$. As this often shifts the solution to a new region of the error space the gradient history for all parameters $p \in x_p$ is reset. Eventually, when a solution is to be returned, the temporary parameters are substituted with their closest variables according to the respective $\mathcal{K}^D_t$. The forward and backward passes are repeated until the error is below the maximum error threshold $e_{max}$ or a maximum number of iterations is reached. After that the optimised program $\rho^*$ is scored according to its error and complexity, and pushed to a priority queue holding potential solutions.

\subsection{Structure search}
We represent the space of possible program structures as a graph $G=(T^{AST}, E)$ where each node $T_i \in T^{AST}$ is a valid program abstract syntax tree (AST). There is an edge from $T_i$ to $T_j$ if and only if $T_j$ can be obtained by replacing exactly one of the leaves in $T_i$ with a subtree $T_s$ of depth 1. The program induction procedure always starts with an empty program. So, we frame structure search as a path finding problem, solved through the use of A* search. 
\paragraph{Score function.}
The total cost function we use is $f_{total}(\rho) = C(\rho) + L(\rho)$, where $L(\rho)$ is the loss function defined in equation (\ref{eq:loss}) and $C(\rho)$ is a function which measures the complexity of the program $\rho$. $C(\rho)$ can be viewed as the cost to reach $\rho$ and $L(\rho)$ as the distance to the desired goal. The complexity function $C(\rho)$ is the weighted sum of (i) maximum depth of the program AST; (ii) the number of free parameters; (iii) the number of variables used by the program; the weights of which we set to $w_C = [10, 5, 1]$. These choices have the effect that short programs maximally exploiting structure from the observation trace are preferred.
\paragraph{Neighbours expansion.}
When the current best candidate solution is popped from the priority queue, we check if it matches the observation trace according to the input specification. If so, the candidate can be returned as the final solution, otherwise it is used as a seed to propose new candidate solutions. Typically in A* search, all neighbouring nodes are expanded and pushed to the priority queue, which is not feasible in our case, though. Therefore, we utilise the available gradients in order to perform a guided proposal selection. Each leaf in the abstract syntax tree $T_\rho$ of a seed candidate solution $\rho$ corresponds to a parameter or a variable. According to the definition of $G$ we need to select exactly $1$ leaf to be replaced with a subtree $T_s$ of depth 1. We select leaf $l \in T_\rho$ according to:
\begin{equation}
l = \argmax{x \in x_p \cup x_v}{\lVert \nabla_x L(\rho) \rVert_2}
\end{equation}
After that, all possible replacement subtrees are constructed. An AST subtree $T_s$ of depth 1 represents a function call. We prune the number of possible functions in $\mathcal{F}$ by ensuring type consistency. Each leaf of $T_s$ can be a parameter or a variable. So, all possible combinations are considered. New variable leaves are initialised to a random variable with suitable type from memory, while new parameter leaves are sampled from the multivariate normal distribution $\mathcal{N}(0, 0.1)$. As a result, if $n_f$ functions are type compatible with $l$ and each function takes $n_a$ arguments at most, then there are $2^{n_a} \cdot n_f$ replacement subtrees, resulting in that many new candidates. All newly proposed candidates are optimised in parallel, scored by $f_{total}$ and pushed to the priority queue.

\section{Experimental results}
In order to evaluate the effectiveness of the $\pi$-machine, we apply it to three different scenarios. Firstly, we consider model learning for a physical system. We show that the $\pi$-machine successfully induces basic physical laws from observational data. Secondly, we demonstrate how a program can be induced in order to explain the behaviour of the policy learnt by a DQN network. This experiment is based on our view that the core deep neural network based policy learner and the explanation layer play complementary roles. As is well known by now, there are numerous advantages to performing end-to-end policy learning, such as DQN-learning from raw video. Having done this, there is also a need to explain the behaviour of the learnt policy with respect to user-defined properties of interest, hence as a program defined in terms of user-defined detectors, $D_v$. Lastly, we apply the $\pi$-machine to a learning-by-demonstration task, where demonstrated behaviour of physical object manipulation is explained through a functional program that describes the abstract structure of that planning task. All experiments are run on an Intel Core i7-4790 processor with 32GB RAM. The $\pi$-machine is implemented in Clojure, which is a LISP dialect supporting powerful data structures and homoiconic syntax. We will be releasing the code once this paper has been accepted for publication. 

Given the nature of the experimental tasks, we use the following list of supported functions, $\mathcal{F}$, for all of our experiments: vector addition, subtraction and scaling. 

\subsection{Physical systems}

\begin{figure}[t]
	\centering
    	\includegraphics[width=0.9\textwidth]{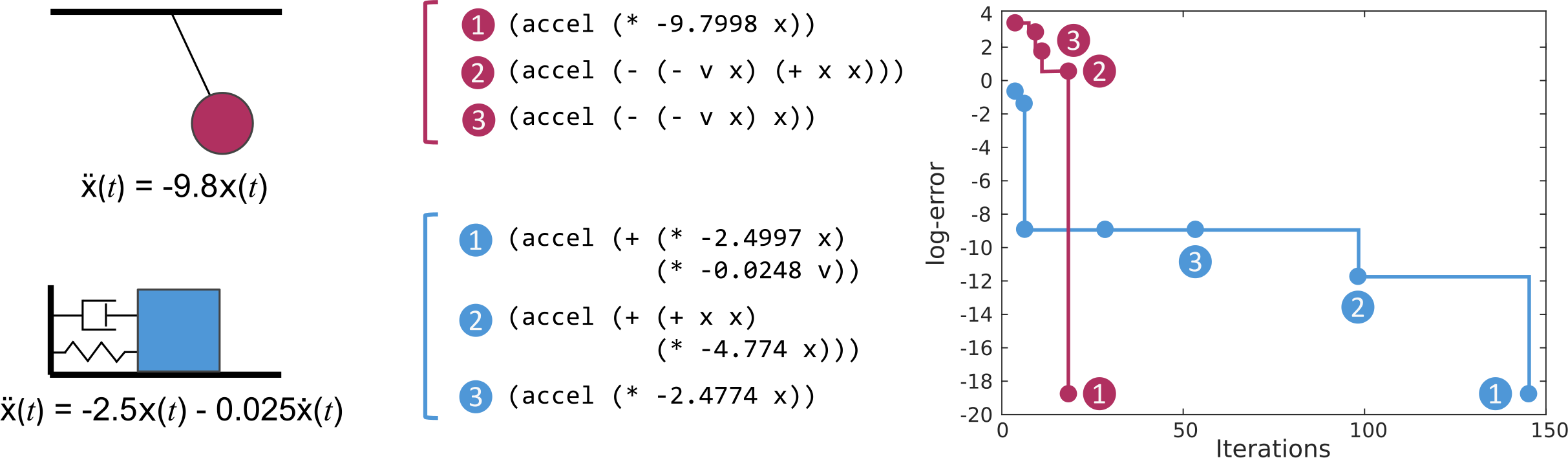}
    \caption{The $\pi$-machine applied to explaining the behaviour of a pendulum (top) and a linear oscillator(bottom). The best 3 solutions for each system are shown in the middle.}
    \label{fig:phy_res}
\end{figure}

The transition dynamics of a second order dynamical system is written as $
\ddot{\mathbf{x}}(t)  = k_1 \mathbf{x}(t) + k_2 \dot{\mathbf{x}}(t)$, where $\mathbf{x}(t)$ is the state of the system at time $t$ and $k_1, k_2$ are system coefficients. We have recreated an experiment described in \citet{schmidt2009}, where the authors show the learning of physical laws associated with classical mechanical systems including the simple pendulum and linear oscillator. A diagram of these two systems is shown in Figure \ref{fig:phy_res} (left). We set $\mathcal{A} = \{accel(\theta)\}$ where $\theta \in \mathbb{R}$ for both experiments. The observation trace for each system is generated by simulating the dynamics for 1s at 100Hz. We specify the action error function as $\sigma_{act} = \lVert \hat{\theta} - \theta \rVert_2$ and set $\sigma_{len} = 0$. In the pendulum experiment, $\mathbf{x} \in \mathbb{R}$ is the angular position of the pendulum, while $\mathbf{v} = \mathbf{\dot{x}} \in \mathbb{R}$ is the angular velocity. In the linear oscillator experiment, $\mathbf{x}$ and $\mathbf{v}$ are the linear position and velocity, respectively.

The three best solutions found by the $\pi$-machine for each system are shown in Figure \ref{fig:phy_res} (middle). The best solution for each system correctly represents the underlying laws of motion. The program describing the behaviour of the pendulum was induced in 18 iterations, while the linear oscillator program needed 146 iterations. The total number of possible programs with AST depth of 2, given the described experimental setup, is approximately $1.7 \times 10^4$. The average duration of an entire iteration (propose new programs, optimise and evaluate) was $0.6s$. \citet{schmidt2009} achieve similar execution times, but distributed over 8 quad core computers (32 cores in total). The experimental results demonstrate that the $\pi$-machine can efficiently induce programs representing fundamental laws of physics.

\subsection{Deep Q-network}

\begin{figure}[t]
	\centering
    	\includegraphics[width=\textwidth]{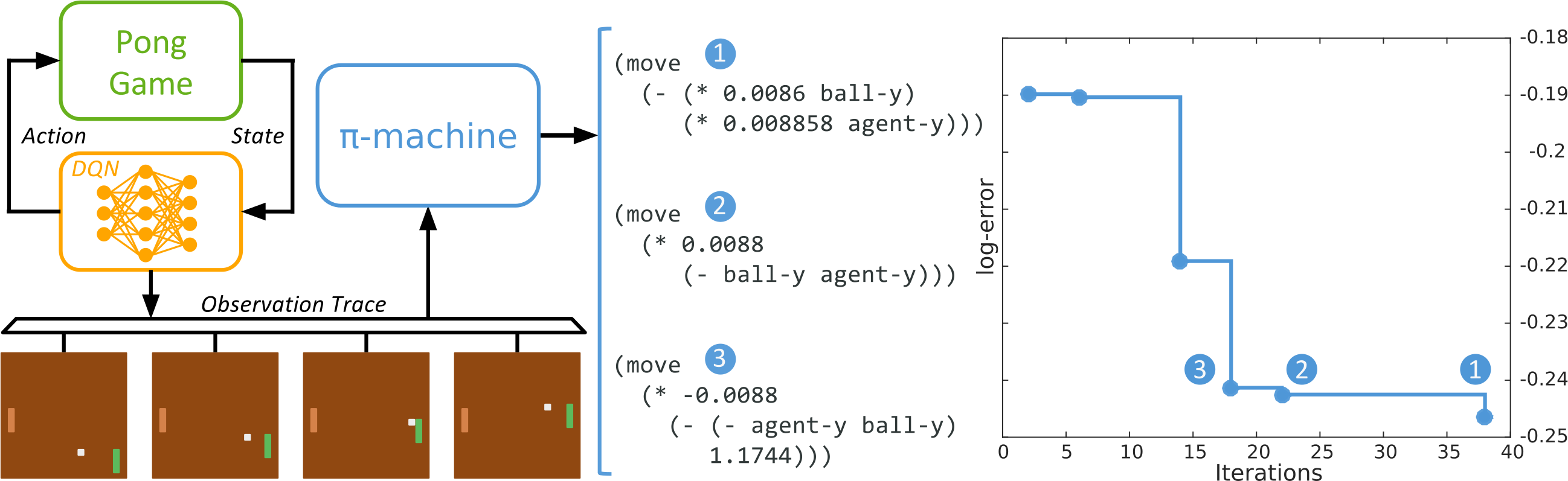}
    \caption{The $\pi$-machine applied to explaining the behaviour of a DQN playing ATARI Pong. The best 3 solutions are shown in the middle.}
    \label{fig:dqn_res}
\end{figure}

We consider explaining the behaviour of a DQN trained to play the ATARI Pong game. We are interested in the question: how does the network control the position of the paddle in order to hit the ball when it is in the right side of the screen. A diagram of the experimental setup is shown in figure \ref{fig:dqn_res} (left). The behaviour of the DQN is observed during a \emph{single} game. Since the environment is deterministic, the state transition function, which generates the observation trace for this experiment, is the policy $\pi(s)$ that the DQN has learnt. We would like to explain the behaviour of the DQN in terms of the position of the opponent, the ball and the DQN agent (so, not just in terms of RAM memory values, for instance). Therefore, the observation trace contains those positions which are extracted from each frame by a predefined detector. We set $\mathcal{A} = \{move(\theta)\}$ where $\theta \in \mathbb{R}$ and represent the discrete actions of the network \verb+left+, \verb+right+, \verb+nop+ as $move(1)$, $move(-1)$, $move(0)$ respectively. We specify the action error function as $\sigma_{act} = \lVert \hat{\theta} - \theta \rVert_2$ and set $\sigma_{len} = 0$. 

The best 3 programs found by the $\pi$-machine are shown in Figure \ref{fig:dqn_res} (middle), where it took 38 iterations for the best one (average iteration duration 3.2s). By inspecting the second solution it becomes clear that the neural network behaviour can be explained as a proportional controller minimising the vertical distance between the agent and the ball. However, the best solution reveals even more structure in the behaviour of the DQN. The coefficient in front of the agent position is slightly larger than the one in front of the ball position which results in a small amount of damping in the motion of the paddle. Thus, it is evident that the DQN not only learns the value of each game state, but also the underlying dynamics of controlling the paddle. Furthermore, we have tested the performance of an agent following a greedy policy defined by the induced program. In our experiments over 100 games this agent achieved a score of $11.1 (\pm 0.17)$. This is not quite the score of $18.9 (\pm 1.3)$ obtained by an optimised DQN, but it is better than human performance $9.3$ \cite{mnih2015}. This difference of course emanates from the predefined detector not capturing all aspects of what the perceptual layers in DQN have learnt, so improved detector choices should yield interpretable programs that also attain performance closer to the higher score of the black-box policy.


\subsection{Learning by demonstration}

\begin{figure}[t]
	\centering
    	\includegraphics[width=\textwidth]{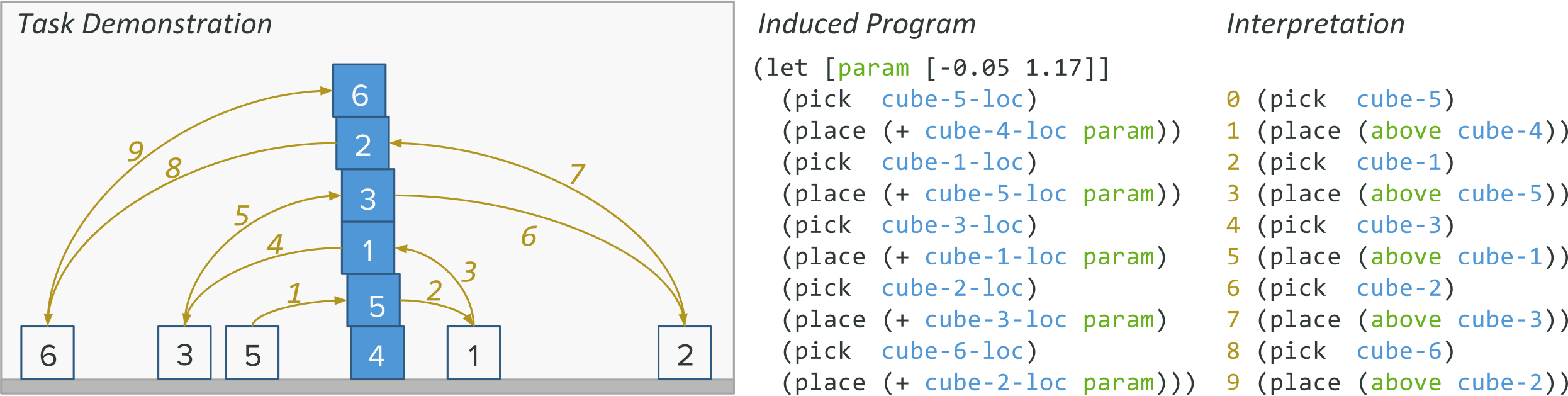}
    \caption{Program induced by the $\pi$-machine explaining a human demonstration of tower building (induced for 27 iterations and 7.2s).}
    \label{fig:hri_res}
\end{figure}

Work in collaborative human-robot interaction \citep{penkov2017} suggests that programmatic description of the task enables robots to better ground symbols to their physical instances, improving their perceptual capabilities. We consider a learning by demonstration scenario where a person demonstrates how to build a tower in a virtual simulated 2D environment. A typical demonstration is depicted in Figure \ref{fig:hri_res} (left). Our goal is to learn a program describing the demonstration such that it could later be utilised by the robot. 

We are interested in how a person moves the cubes through the entire demonstration. So, we set $\mathcal{A} = \{pick(\theta), place(\theta)\}$, where $\theta \in \mathbb{R}^2$ is a 2D location. We specify the action and length error functions as
\begin{equation}
\sigma_{act}(\hat{a}, \hat{\theta}, a, \theta) = 
	\begin{cases}
    \lVert \hat{\theta} - \theta \rVert_2 & \text{if } \hat{a} = a \\
    d_{max} & \text{otherwise}
	\end{cases}
\qquad\qquad
\sigma_{len}(T', T) =  (T' - T)^2
\end{equation}
where $d_{max}$ is the maximum distance within the simulated 2D environment. The states in the observation trace contain the 2D position of every cube in the environment. 

We tested the $\pi$-machine on 300 demonstrations and it successfully induced a program for each one of them. On average, 67 iterations (average iteration duration 0.4s) were needed per demonstration. The solution for one of the demonstrations is shown in Figure \ref{fig:hri_res}. From inspection, it can be seen that the program not only describes which cube was picked up and put where, but also what the spatial relations are when stacked. The $\pi$-machine induced and optimised a free parameter which is used to encode the relation ``above'' as \verb!(+ loc [-0.05 1.17])!. This type of information could speed up task learning and improve the robot's ability to understand its environment.

\section{Discussion}

The $\pi$-machine can be viewed as a framework for automatic network architecture design \citep{zoph2017,negrinho2017}, as different models can be expressed as concise LISP-like programs (see Figure \ref{fig:equivalence}). Deep learning methods for limiting the search space of possible programs, which poses the greatest challenge, have been proposed \citep{balog2016}, but how they can be applied to more generic frameworks such as the $\pi$-machine is an open question. The specification of variable detectors not only addresses this issue, but enables the user to make targeted and well grounded queries about the observed data trace. Such detectors can also be learnt from raw data in an unsupervised fashion \citep{garnelo2016,kim2017}.


\section{Conclusion}
We propose a novel architecture, the $\pi$-machine, for inducing LISP-like functional programs from observed data traces by utilising backpropagation, stochastic gradient descent and A* search. The experimental results demonstrate that the $\pi$-machine can efficiently and successfully induce interpretable programs from short data traces. 
\bibliography{references}
\end{document}